\documentclass{article}
\usepackage{amsmath,epsfig}
\usepackage{graphicx}
\usepackage{subfigure}
\usepackage{fancyhdr}

\usepackage[preprint]{spconfa4}
\copyrightnotice{978-1-6654-3864-3/21/\$31.00  \copyright 2021 IEEE}

\let\OLDthebibliography\thebibliography
\renewcommand\thebibliography[1]{
  \OLDthebibliography{#1}
  \setlength{\parskip}{0pt}
  \setlength{\itemsep}{0pt plus 0.3ex}
}

\usepackage{enumitem}
\usepackage{pifont}
\usepackage[usenames,dvipsnames]{xcolor}
\definecolor{dgreen}{rgb}{0.0,0.6,0.0}
\newcommand{\cmark}{\textcolor{dgreen}{\text{\ding{51}}}}%
\usepackage[pagebackref=true,breaklinks=true,colorlinks,bookmarks=false]{hyperref}

\fancypagestyle{FirstPage}{
	
	\lfoot{978-1-6654-3864-3/21/\$31.00  \copyright 2021 IEEE}
	
}

\begin{document}\sloppy
	
	
	\def\eg{\emph{e.g}\onedot} \def\Eg{\emph{E.g}\onedot}
	\def\ie{\textit{i.e}} \def\Ie{\emph{I.e}\onedot}
	\def\etc{\emph{etc}\onedot}

	\def\x{{\mathbf x}}
	\def\L{{\cal L}}

	\title{Input-Output Balanced Framework for Long-tailed LiDAR Semantic Segmentation
	}
	%
	\name{Peishan Cong$^{\ast}$, Xinge Zhu$^{\dagger}$, Yuexin Ma$^{\ast,\ddagger}$}
	\address{$^{\ast}$ ShanghaiTech University $^{\dagger}$ Chinese University of Hong Kong \\$^{\ddagger}$ Shanghai Engineering Research Center of Intelligent Vision and Imaging\\ congpsh@shanghaitech.edu.cn \quad mayuexin@shanghaitech.edu.cn}

	\maketitle

\begin{abstract}
	
	A thorough and holistic scene understanding is crucial for autonomous vehicles, where LiDAR semantic segmentation plays an indispensable role. However, most existing methods focus on the network design while neglecting the inherent difficulty, \ie, imbalanced data distribution in the realistic dataset (also named long-tailed distribution), which narrows down the capability of state-of-the-art methods. In this paper, we propose an input-output balanced framework to handle the issue of long-tailed distribution. Specifically, for the input space, we synthesize these tailed instances from mesh models and well simulate the position and density distribution of LiDAR scan, which enhances the input data balance and improves the data diversity. For the output space, a multi-head block is proposed to group different categories based on their shapes and instance amounts, which alleviates the biased representation of dominating category during the feature learning. We evaluate the proposed model on two large-scale datasets, \ie, SemanticKITTI and nuScenes, where state-of-the-art results demonstrate its effectiveness. The proposed new modules can also be used as a plug-and-play, and we apply them on various backbones and datasets, showing its good generalization ability.
	
	
\end{abstract}

\section{Introduction}
\thispagestyle{FirstPage}

\begin{figure}[t]
	\includegraphics[width=0.98\columnwidth]{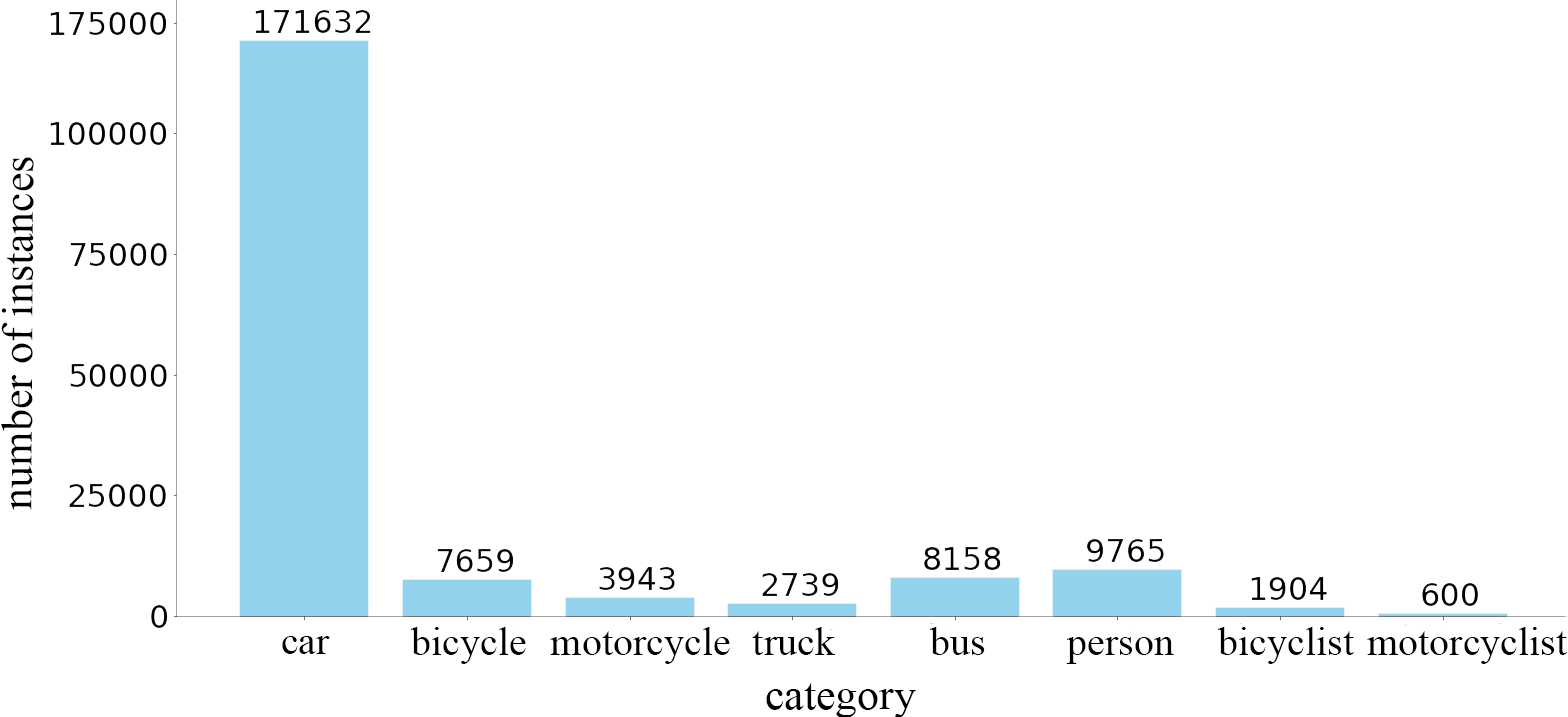}
	\vspace{-2ex}
	\caption{The statistic of number of different categories in the SemanticKITTI~\cite{behley2019semantickitti} dataset.}
	\label{fig:occur_SemanticKITTI}
	\vspace{-1.5ex}
\end{figure}

Autonomous vehicle, one of the most promising and popular applications of computer vision, has witnessed great progress in recent years, in which 3D LiDAR sensor plays a key role in the scene perception because it provides more precise and farther-away distance measurements of the surrounding environments than  ordinary visual cameras. LiDAR semantic segmentation that could provide a thorough scene understanding, has attracted extensive studies. Most existing works usually pay much attention on point-to-structure representations, including spherical projection ~\cite{xu2020squeezesegv3,milioto2019rangenet++}, bird-eye-view projection~\cite{Zhang2020PolarNetAI} and 3D voxelization~\cite{Zhu2020CylindricalAA,Hong2020LiDARbasedPS}, and another group of methods focus on the network architecture design~\cite{wu2019squeezesegv2,Zhu2020CylindricalAA,Zhou2020Cylinder3DAE}. However, these methods often neglect the inherit difficulty caused by the long-tailed data distribution. Specifically, in the realistic dataset, the data distribution is usually imbalanced. We show a statistic of data distribution on SemanticKITTI dataset in Fig.~\ref{fig:occur_SemanticKITTI}. It can be found that most categories, including bicycle, motorcycle, truck,etc. only cover less than 20\% while the car category is in the dominating level and makes up about 80\% of whole data. The extreme long-tailed distribution can be easily observed in the realistic dataset, and this bias would make these tailed categories be overwhelmed by the majority categories during training and make the feature representation incline to the dominating categories.

In this work, we focus on the difficulty of imbalanced data distribution and introduce an input-output balanced framework to perform long-tailed semantic segmentation on 3D LiDAR point cloud in the input and output space, respectively. 
For the input level, we first analyse the principle of LiDAR emission and the density and position distribution of original database. Then we employ the mesh models from different tail categories to simulate the realistic instance point cloud and an instance-level database is created to store these tail categories. In this way, the generated points suit the distribution of objects in realistic dataset and also improve the data diversity because these mesh models provide different poses and positions . During training, we introduce several samples from the database to balance the category distribution. For the output level, we design a multi-head block to group these categories based on the shape and instance amounts, which alleviates the interference of dominating category and helps the model learn better representations for tails. Extensive experiments have been conducted on two large scale datasets, including SematicKITTI~\cite{behley2019semantickitti} and nuScene~\cite{caesar2020nuscenes} and experimental results demonstrate the effectiveness of our method. The proposed two modules can also be used as a plug-and-play, and we apply them on various backbones and datasets, showing its good generalization ability.

Our contribution can be summarized as follows.\begin{enumerate}
	\item[(1)] We propose an input-output balanced framework to tackle the issue of long-tailed data distribution in the LiDAR semantic segmentation. 
	\vspace{-1.5ex}
	\item[(2)] We employ mesh models to generate a database consisting the tail objects, which simulates the position and density distribution in the original dataset. Moreover, the generated instance-level point cloud could better enhance the data diversity. 
	\vspace{-1.5ex}
	\item[(3)] We design an output balanced module using a combination of multi-head that groups the category based on the shape and instance amounts. The operation of dividing different category apart alleviates the interference of dominating category and promote the model learn better representations for tails.
\end{enumerate}

\section{Related work}
\subsection{LiDAR point cloud segmentation}
Different from indoor-scene point cloud segmentation~~\cite{Qi2017PointNetDL,Thomas2019KPConvFA,Wang2019DynamicGC}, outdoor LiDAR point cloud segmentation has much more challenges because of the varying density of points, the large number of points, and the large range of scenes. Considering the sparsity of LiDAR point cloud and the computational complexity, most methods transform 3D point cloud to 2D grids and adopt 2D CNN for segmentation. SqueezeSeg~\cite{wu2018squeezeseg}, SqueezeSegv2~\cite{wu2019squeezesegv2}, SqueezeSegv3~\cite{xu2020squeezesegv3}, Darknet~\cite{behley2019semantickitti}, and RangeNet++~\cite{milioto2019rangenet++} utilize the spherical projection to convert the point cloud to a range image. These methods inevitably brings discretization errors and occlusions and makes long-tailed problem more serious since some categories with fewer points may be sheltered by others. PolarNet~\cite{Zhang2020PolarNetAI} introduced bird’s-eye-view representation under the polar coordinates to balance the points across grid cells in a polar system. To maintain the 3D geometric relation of points, some methods~\cite{Graham20183DSS,Han2020OccuSegO3,Tchapmi2017SEGCloudSS,Wang2020ReconfigurableVA,Zhu2020SSNSS} use the voxel partition and apply regular 3D convolutions for LiDAR segmentation. But their performances become limited for outdoor scenes due to the special properties of outdoor LiDAR point cloud mentioned above. Then, focusing on the characteristics of sparsity and
varying density,~\cite{Zhu2020CylindricalAA} proposed a cylindrical and asymmetrical 3D convolution networks for LiDAR segmentation and achieved state-of-the-art performance. However, all these methods still leave the long-tailed problems, where the accuracy of rare categories are no less unsatisfying. 
\subsection{Long-tailed perception methods}
Long-tailed problem exists in many perception tasks, like the object detection and instance segmentation. There are three main streams for alleviating the category-unbalanced problem brought by long-tailed data distribution, including re-sampling, re-weighting, and feature manipulation methods. Re-sampling methods~\cite{hu2020learning,han2005borderline,long-tailed} increase the samples of minority classes while undersample those from frequent classes in training. ~\cite{yan2018second} utilizes the re-sampling method in point cloud detection by oversampling rare-category objects with the bounding box in ground truth. However, for semantic segmentation task, there are no bounding box or annotation for instances, which makes it impossible to do re-sampling operations from the original dataset. In addition, in large-scale outdoor scenes, instances have various movements and distributions, resulting in heterogeneous orientations, distances, and number of LiDAR points. The original dataset lacks of the diversity and re-sampling can cause over-fitting problems due to the few samples. Our method simulates the point cloud for instances with authenticity and diversity and overcomes the limitation of above methods.

Based on the inverse of class frequency or a function of class frequency~\cite{imbalanced}, re-weighting methods ~\cite{cui2019class,li2020overcoming} assign different weights for different training samples so that the long-tailed effect can to some extend be counteracted by assigning a lower weight to the head category and a higher weight to the tail category.
There are many other methods focusing on the feature representation by including a hierarchical tree structure for better classification~\cite{ouyang2016factors} or transferring the feature variance of the regular class with sufficient training samples to increase the feature space of the tail class~\cite{yin2019feature}. Our proposed model is complementary for these methods and can be combined together to further improve the performance.


\section{ Methodology}
\begin{figure*}[t]
	\centering
	\includegraphics[width=0.9\linewidth]{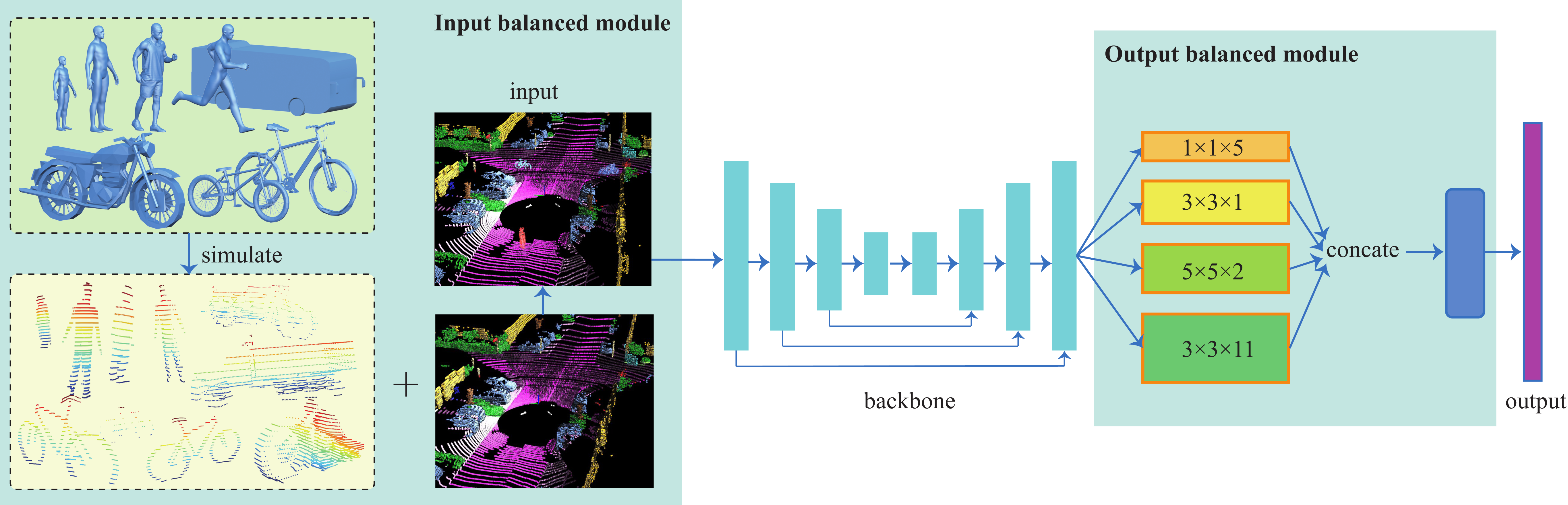}
	\vspace{-1ex}
	\caption{Our Input-output balanced Framework. It mainly contains two modules. The input balanced module simulates LiDAR point cloud for tailed categories by 3D mesh models and LiDAR working principles and we add these samples into the original dataset to balance the distribution during training. The output balanced module utilizes different heads for different groups of categories to alleviate the biased representation of dominating categories. }
	\label{fig:framework}
\end{figure*}
\subsection{Overview}
For long-tailed datasets, all the positive samples of other categories can become negative samples for the target category. The tail categories are easily overwhelmed by major categories. The key problems are the unbalanced data and the biased training procedure. Our approach tackles these key problems in the LiDAR point cloud segmentation by proposing the input balanced module and the output balanced module. The whole pipeline of our model is shown in Fig.~\ref{fig:framework}, which consists of two major components (in the green rectangles). Note that the backbone in our method is flexible and we apply various backbone networks in the experiments.


\subsection{Input Balanced Model}
\subsubsection{Data Distribution}
\begin{figure}[t]
	\subfigure[pedestrian distribution]{
		\includegraphics[width=0.45\linewidth]{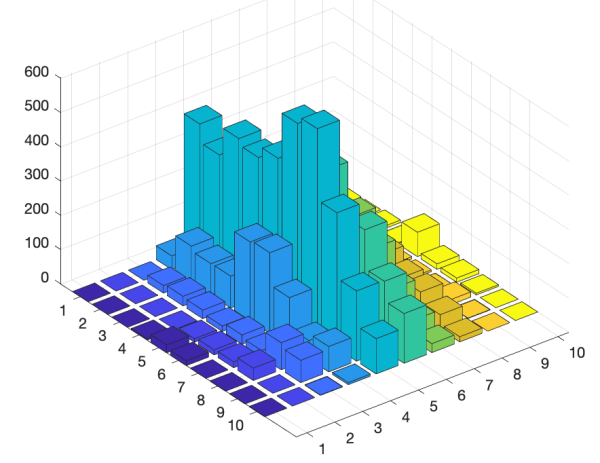}
	}
	\subfigure[truck distribution]{
		\includegraphics[width=0.45\linewidth]{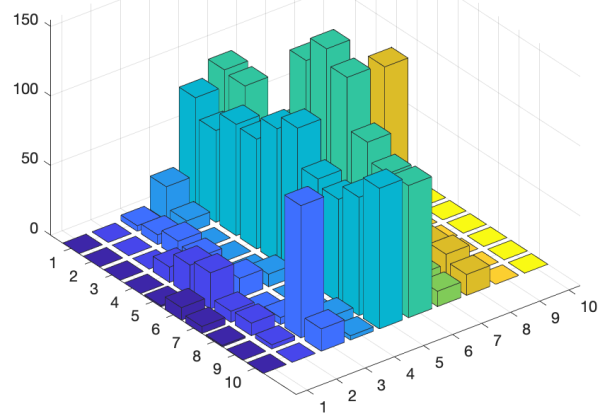}}
	\vspace{-2ex}
	\caption{Density and position distribution of instances on training set in semantic KITTI. We divide the area of LiDAR point cloud into grids with each size is $10 \times 10$ (the unit is meter) according to xy-coordinate and record the occurrence of objects in different grids, and the height of the cube represents the occurrence number. }
	\label{fig:truck-distribution}
\end{figure}

To better simulate various objects and benefit the training process, it is essential to follow the distribution of objects in the original dataset for autonomous driving. During data collection, the acquisition car is in the center with a LiDAR sensor on the top of the car. The point cloud data is scanned by LiDAR in $360^\circ$ field of the horizontal view and a fixed range
in the vertical view. There are a variety of road conditions in reality including straight road, crossroads and double direction path, and vehicles always appear on the road or parking area. Due to the variety of road conditions and road distribution, objects in dataset are not uniformly distributed. To model the object distribution, we divide the area of the captured point cloud into $10\times 10$ voxels along x and y axis and calculate the number of instances per category in each voxels. 
The object distribution of different categories in SemanticKITTI is shown in Fig.~\ref{fig:truck-distribution} (take truck and person as example). 

\subsubsection{Database}
Based on the position and density distribution of each tailed category, we create a large-scale database. We visualize parts of the pedestrian database in Fig.~\ref{fig:database}. There are two main steps of the database construction: choosing relative positions for instances to the LiDAR sensor and calculating the simulated point cloud.
In this section, we introduce the details of the database construction.

\begin{figure}
	\centering
	\includegraphics[width=0.85\linewidth]{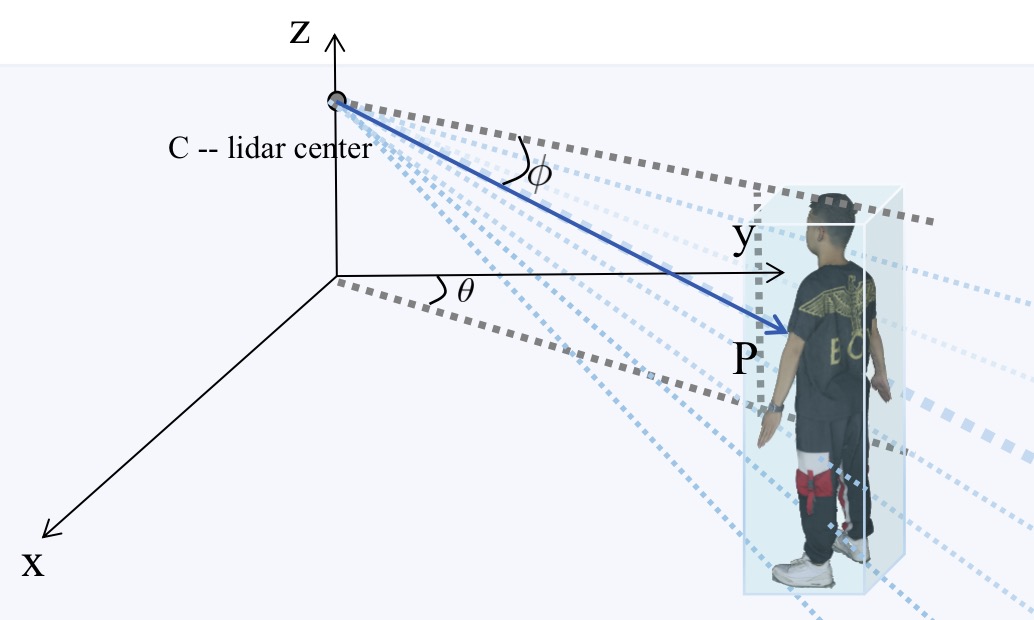}
	\vspace{-1ex}
	\caption{The calculation of intersection points between LiDAR emissions and the mesh model. The blue lines represent LiDAR emitting directions, one of these direction in solid line gives the intersection point $P$ on one mesh face of the model.}
	\label{fig:lidar}
\end{figure}

The emitted light from LiDAR will bounce back when hitting an obstacle, and the generated points are in relative positions of LiDAR. In addition, the traffic participants, like the vehicles, bicycles, and pedestrians, are inner categories with similar shape. We can use diverse 3D mesh models to cover most of the cases of traffic participants. Thus, according to this LiDAR working principle, we simulate the point cloud of these tail categories via 3D mesh models. The HDL-64E LiDAR sensor is designed in $360^\circ$ field of horizontal view with $0.08^\circ$ angular resolution and in $-27^\circ \sim 0^\circ$ vertical view with $0.4^\circ$ angular resolution. In order to completely simulate the point cloud generated by LiDAR, we employ the direction of LiDAR light emission with above ranges and resolutions. We describe each emission with a unit vector in spherical coordinates as $t = [\cos\phi\sin\theta,\cos\phi\cos\theta,\sin \phi] $, where $\phi$ denotes the angle between the emitting direction and plane xy and $\theta$ is the azimuth. These two angles are marked in Fig ~\ref{fig:lidar}. The LiDAR center C is [0,0,2], because the moving car is in the center and LiDAR sensor is on the top of car with z coordinate is about 2. The model placing position is randomly chosen from object distribution in Fig ~\ref{fig:truck-distribution}, relative size of different objects is also imitated as that in the original dataset. 

Moreover, we employ a random rotation and flip on mesh model and calculate the LiDAR scanning range of angular $[\theta_1,\theta_2]$ and vertical $[\phi_1,\phi_2]$ according to the object bounding box to speed up the computation. In the real world, the emissions are not completely evenly distributed, so we also add noise on the emission distribution.

\begin{figure}
	\subfigure[pedestrian in the original dataset]{
		\includegraphics[width=0.46\columnwidth,height=0.3\columnwidth]{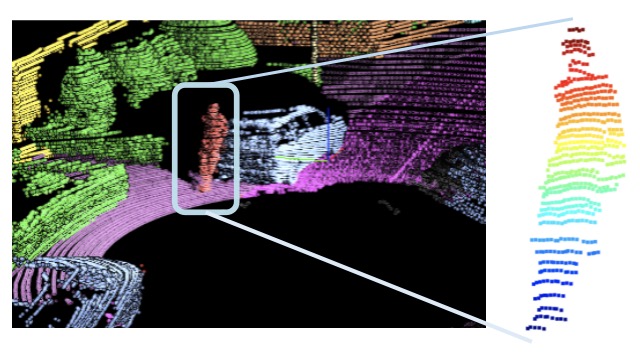}
		\label{fig:image-2}
	}
	\subfigure[simulated pedestrian]{
		\includegraphics[width=0.46\columnwidth,height=0.3\columnwidth]{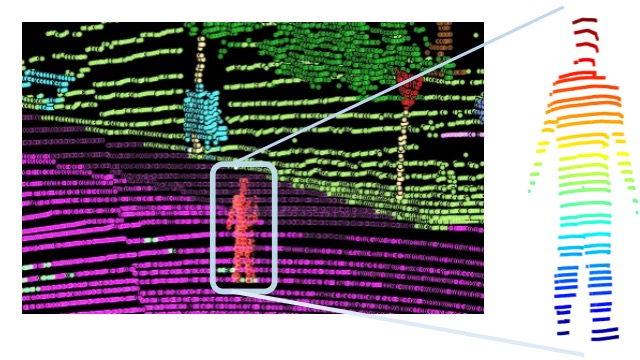}
		\label{fig:image-3}
	}
	\vspace{-2ex}
	\hspace*{-0.2cm} 
	\subfigure[simulated pedestrian near the LiDAR center]{
		\includegraphics[width=0.45\columnwidth,height=0.3\columnwidth]{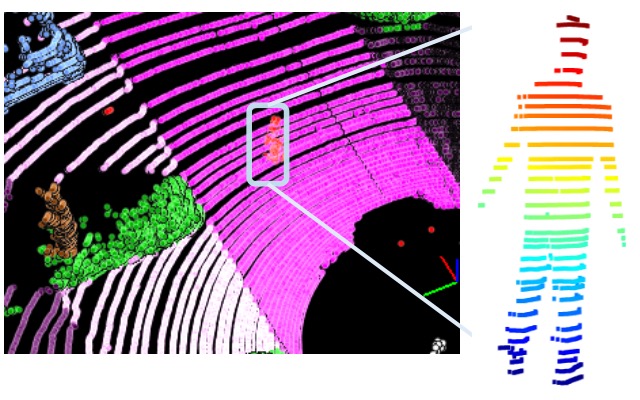}
		\label{fig:image-dis}
	}
	\hspace*{0.2cm} 
	\subfigure[simulated pedestrian far from the LiDAR center]{
		\includegraphics[width=0.45\columnwidth,height=0.3\columnwidth]{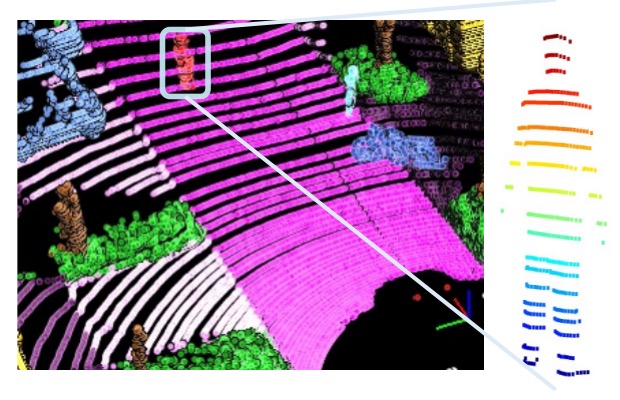}
		\label{fig:image-dis2}
	}
	\vspace{-2ex}
	\hspace*{-0.1cm} 
	\subfigure[simulated bicycle ]{
		\includegraphics[width=0.45\columnwidth,height=0.3\columnwidth]{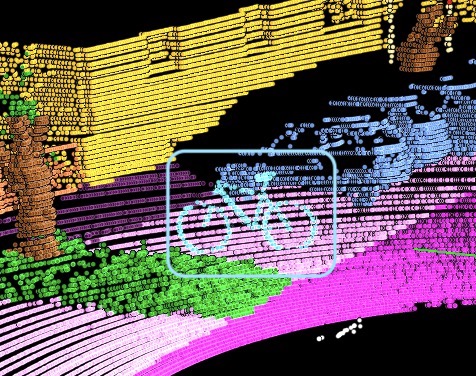}
		\label{fig:image-4}
	}
	\hspace*{0cm} 
	\subfigure[simulated different categories]{
		\includegraphics[width=0.45\columnwidth,height=0.3\columnwidth]{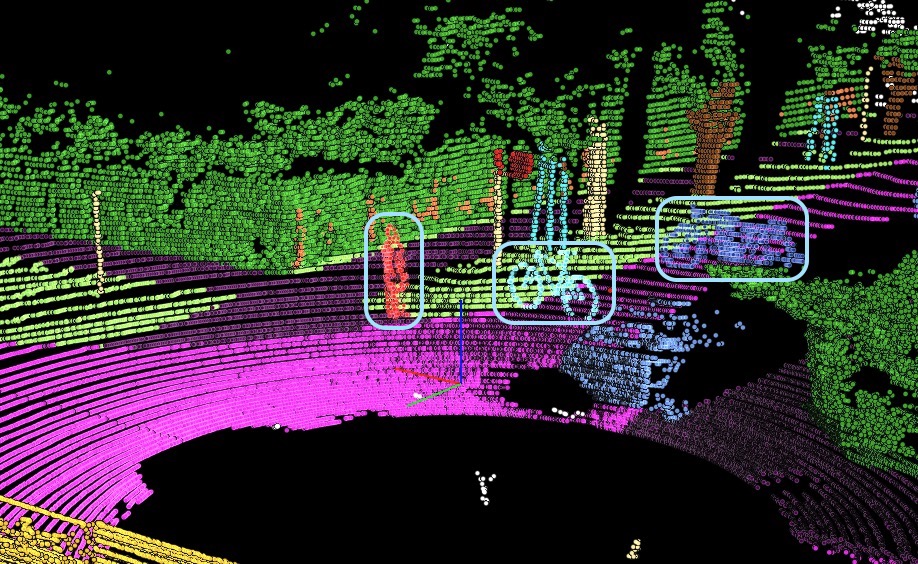}
		\label{fig:image-5}
	}
	
	\caption{The visualization of our generated instances in the dataset. 
	}
	\label{fig:data_distribution}
\end{figure}

The intersection point $p = [p_x,p_y,p_z]$ on mesh in Fig.~\ref{fig:lidar} is calculated in the formula below:\\
\begin{equation}
\mathbf{p} = \mathbf{c} + \mathbf{t}\dfrac{\mathbf{n}^T(q-c)}{\mathbf{n}^T \mathbf{t}},
\end{equation}
where $\mathbf{n}$ represents the normal vector of corresponding mesh and $q$ denotes any vertex point on this mesh, $c$ is the LiDAR center position and $\mathbf{t}$ is the direction vector emitted from LiDAR center in Cartesian coordinate. We also check whether the intersection point is on the model by the following rule:
\begin{figure}
	\centering
	\includegraphics[width=0.9\columnwidth,height=0.3\columnwidth]{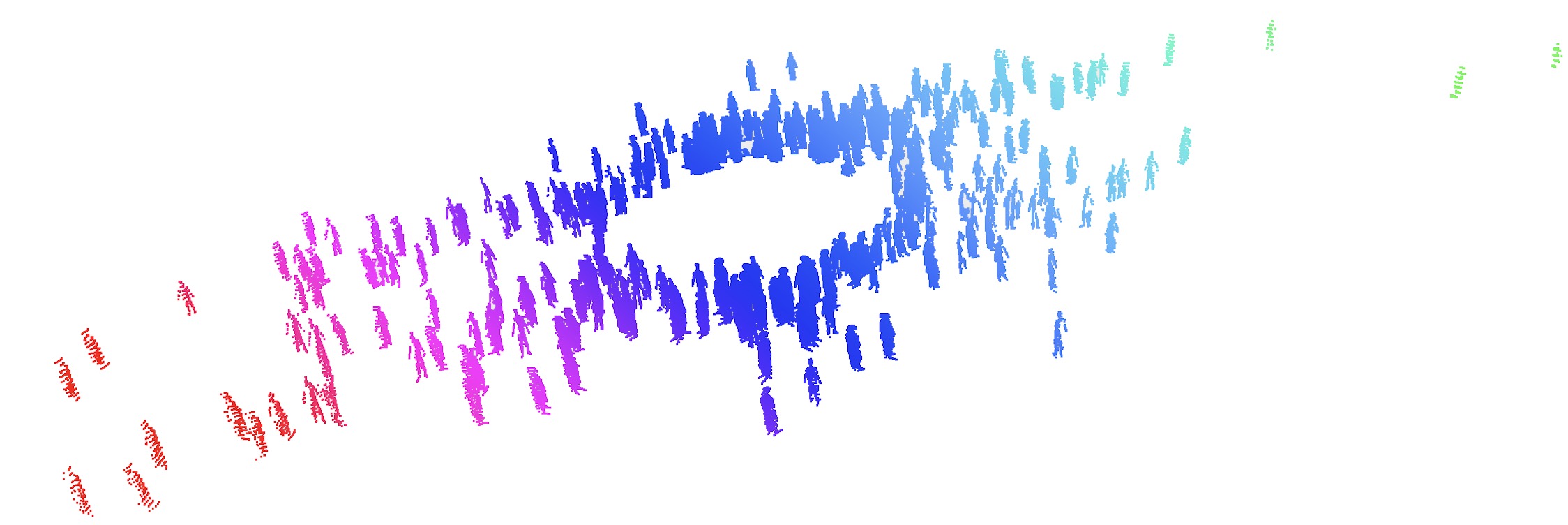}
	\vspace{-1ex}
	\caption{The database for pedestrians. For clear visualization, we illustrate parts of the database. The distribution of instances in the database is based on the statistics in Fig.~\ref{fig:truck-distribution}. The complete database is diverse in the positions, the rotations, the actions, and the scale of the models.
	}
	\label{fig:database}
\end{figure}
\begin{equation}
\begin{aligned}
& \mathbf{p} = 2\mathbf{q} + u(\mathbf{m}-\mathbf{q}) + v(\mathbf{n}-\mathbf{q}), \\
&s.t. \quad u > 0,v>0,u+v < 1,
\end{aligned}
\end{equation}
where $q,m,n$ denotes the vertex coordinates of the mesh face and  $u,v$ is the constrain of a linear combination. The position choosing is under Cartesian coordinate and we transfer it into polar coordinate during training. Fig \ref{fig:image-2} and \ref{fig:image-3} illustrates that our simulated model is similar with the instance from original dataset. We also consider the principle that the LiDAR light will bounce back when hitting an obstacle, resulting in partial observation. In addition, points become sparser when instance is farther away from the LiDAR center due to the attenuation of the signal. The generated model in our method meets all the rules, which can be seen in Fig \ref{fig:data_distribution}. According to different types of LiDARs and traffic scenarios in different datasets, our method can easily generate corresponding database by adjusting the parameters.
\subsubsection{Data Sampling}
We have generated an offline database containing the labels of all ground truths and their associated point cloud data, the next step is randomly selecting several instances from this database and merge them into the current training point cloud. In our experiments, we add one sample per category in one scan. Moreover, we perform a fast collision test when merging the objects into the dataset to avoid impossible overlapping and physically collision. 
Through this approach, we greatly increase the number of such long-tailed samples per point cloud scan and follow the object distribution, which solves the unbalanced category distribution problem from the input level.

\begin{table*}[ht!]\tiny
	\centering
	\caption{Segmentation results on test split of SemanticKITTI.} \label{tab:test}
	\setlength{\tabcolsep}{2.22mm}
	\begin{tabular}{c|c|c|c|c|c|c|c|c|c|c|c|c|c|c|c|c|c|c|c|c}
		\hline
		{method}&\bf{mIOU}&	\rotatebox{90}{car}&\rotatebox{90}{bicycle}&\rotatebox{90}{motorcycle}&\rotatebox{90}{truck}&\rotatebox{90}{bus}&\rotatebox{90}{person}&\rotatebox{90}{bicyclist}&\rotatebox{90}{motorcyclist}&\rotatebox{90}{road}&\rotatebox{90}{parking}&\rotatebox{90}{sidewalk}&\rotatebox{90}{other-ground}&\rotatebox{90}{building}&\rotatebox{90}{fence}&\rotatebox{90}{vegetation}&\rotatebox{90}{trunk}&\rotatebox{90}{terrain}&\rotatebox{90}{pole}&\rotatebox{90}{traffic-sign} \\
		\hline\hline
		TangentConv~\cite{tatarchenko2018tangent} &35.9&86.8&1.3&12.7&11.6&10.2&17.1&20.2&0.5&82.9&15.2&61.7&9.0&82.8&44.2&75.5&42.5&55.5&30.2&22.2\\
		\hline
		Darknet53 (ICCV2019)~\cite{behley2019semantickitti}&49.9&86.4&24.5&32.7&25.5&22.6&36.2&33.6&4.7&91.8&64.8&74.6&\bf{27.9}&84.1&55.0&78.3&50.1&64.0&38.9&52.2\\
		\hline
		RandLA-Net (CVPR2020) ~\cite{hu2020randla}&50.3&94.0&19.8&21.4&\bf{42.7}&\bf{38.7}&47.5&\bf{48.8}&4.6&90.4&56.9&67.9&15.5&81.1&49.7&78.3&60.3&59.0&44.2&38.1\\
		\hline
		RangeNet++ (IROS2019)~\cite{milioto2019rangenet++}&52.2&91.4&25.7&34.4&25.7&23.0&38.3&38.8&4.8&\bf{91.8}&\bf{65.0}&75.2&27.8&87.4&58.6&80.5&55.1&64.6&47.9&55.9\\
		\hline
		PolarNet (CVPR2020)~\cite{Zhang2020PolarNetAI}&54.3&93.8&40.3&30.1&22.9&28.5&43.2&40.2&5.6&90.8&61.7&74.4&21.7&90.0&61.3&84.0&65.5&67.8&51.8&57.5\\
		\hline
		SqueezeSegv3 (ECCV2020)~\cite{xu2020squeezesegv3}&55.9&92.5&38.7&36.5&29.6&33.0&45.6&46.2&\bf{20.1}&91.7&63.4&74.8&26.4&89.0&59.4&82.0&58.7&65.4&49.6&58.9\\
		\hline\hline
		Ours&\bf{58.2}&\bf{95.0}&\bf{48.4}&\bf{39.9}&35.4&33.1&\bf{52.6}&42.5&13.2&91.6&64.3&\bf{75.2}&20.8&\bf{91.2}&\bf{63.8}&\bf{84.9}&\bf{68.3}&\bf{68.8}&\bf{57.2}&\bf{59.4}\\
		\hline
	\end{tabular}
\end{table*}

\subsection{Output Balanced Module}
To reduce the impact of dominating categories on tail classes, we design an output balanced module before classifier, as shown in Fig.~\ref{fig:framework}. This multi-head block is a combination of a set of convolutions having the kernel size of 1,3, and 5, which separates the feature maps from various categories according to instance amounts (occurrence frequencies) and shapes. In this way, these dividing heads can alleviate the inter-class competition between dominating categories and rare classes, thus reducing the effect of the overwhelmed discouraging signals on tail classes.
Moreover, it is better to enlarge the receptive fields for larger objects such as trucks and buses to capture more descriptive spatial features, and use a smaller kernel size ($1 \times 1$ convolution) for smaller objects such as pedestrian, cyclist and motorcyclist.  
Compared to simple convolution, it helps model learn better representations for tail classes by alleviating the biased representation of dominating categories.

\begin{table}[ht]\tiny
	\caption{Segmentation results on validation split of nuScene.} \label{tab:test on nu}
	\setlength{\tabcolsep}{3.46mm}
	\begin{tabular}{c|c|c|c|c|c}
		\hline
		method & \textbf{mIOU} & \rotatebox{90}{bicycle} & \rotatebox{90}{motorcycle} & \rotatebox{90}{traffic\_cone}& \rotatebox{90}{trailer} \\
		\hline\hline
		PolarNet (CVPR2020)~\cite{Zhang2020PolarNetAI} & 69.962   & 35.92 & 72.39 & 54.66 & 56.63 \\
		\hline
		$+$ Input Balanced Module & 71.416  & 40.72 & 71.66 & \bf{59.29} & 60.55 \\
		\hline
		$+$ Output Balanced Module & 71.462 &  40.18 & 70.52 & 58.73 & 58.51\\
		\hline
		\hline
		Input-Output Balanced Model & \bf{72.256} & \bf{42.15} & \bf{74.29} & 58.47 & \bf{63.24} \\
		\hline
	\end{tabular}
\end{table}

\section{Experiment}

\begin{table}[h]\small
	\centering
	\setlength{\tabcolsep}{0.62mm}
	\caption{Ablation studies for input-output components on SemanticKITTI validation set.} \label{tab:Ablation1}
	\begin{tabular}{ccc|c}
		\hline
		PolarNet & Input Balanced Module & Output Balanced Module & mIOU \\
		\hline 
		\cmark& & & 56.46\\	
		\cmark&\cmark& & 58.137\\	
		\cmark& &\cmark & 	57.837\\ 
		\cmark& \cmark& \cmark& 58.324\\	\hline 
	\end{tabular}
	\vspace{-4ex}
\end{table}

\begin{table}[h]\small
	\centering
	\caption{Ablation studies for input balanced module on different backbones on SemanticKITTI validation set.} \label{tab:Ablation2}
	\begin{tabular}{c|c}
		\hline
		method&mIOU\\
		\hline 	\hline 
		PolarNet & 56.46\\	\hline
		PolarNet + Input Balanced Module &  58.137 \\	\hline
		Salsanext & 57.547 \\	\hline
		Salsanext + Input Balanced Module &  58.452 \\	\hline
	\end{tabular}
\end{table}

In this section, we first describe our experimental setup, and then conduct the proposed method on two large scale datasets. 
Ablation studies are also conducted to validate each component of our approach and its generalization ability.
\subsection{Datasets and Evaluation Metric}
\noindent\textbf{SemanticKITTI} It contains 21 sequences of point-cloud data with 43,442 densely annotated scans and total 4549 millions points. Over 21K scans (sequences between 00 and 10) are used for training, where scans from sequence 08 are particularly used in validation. The remaining scans (between sequences 11 and 21) are used as test split. 19 categories are remained after ignoring classes with very few points.

\noindent\textbf{nuScenes} It collects 1000 scenes of 20s duration with 32 beams LiDAR sensor. The total frames is sampled at 20Hz and the number is 40,000 with officially splitting as training and validation set. The number of the categories after merging similar classes and removing rare classed is 16. 

\noindent\textbf{Evaluation Metric} 
To evaluate the results of our model, we follow previous work~\cite{Zhu2020CylindricalAA,Zhang2020PolarNetAI} and use mean intersection-over-union (mIoU) over all classes as the evaluation metric, which is given by
$
m I o U=\frac{1}{C} \sum_{i=1}^{C} \frac{\left|\mathcal{P}_{i} \cap \mathcal{G}_{i}\right|}{\left|\mathcal{P}_{i} \cup \mathcal{G}_{i}\right|},
$
where $\mathcal{P}_{i}$ and $\mathcal{G}_{i}$ denote the predicted and labelled set of points for class $i$, and $C$ is the number of total categories.

\subsection{Results on SemanticKITTI}
In this experiment, we compare the results of our pro-posed method with existing state-of-the-art LiDAR segmentation methods on SemanticKITTI test set. In the implementation, we use the PolarNet~\cite{Zhang2020PolarNetAI} as our backbone, and employ the proposed two modules based on this structure. 
As shown in Table.~\ref{tab:test}, our method achieves state-of-the-art performance compared to these existing models~\cite{xu2020squeezesegv3,hu2020randla,Zhang2020PolarNetAI}. Especially, for these tailed categories, including bicycle, motorcycle, person, the proposed method achieves the better performance with a clear margin, which demonstrates the effectiveness of the proposed input-output balanced modules.

\subsection{Results on nuScenes}
Since the nuScenes dataset is a latest released dataset, there is no existing work conducted on it. We implement the PolarNet~\cite{Zhang2020PolarNetAI} by ourselves, and add the proposed modules, input balanced module and output balanced module, respectively, to investigate the effectiveness of these two components. We add the new classes traffic cone and trailer in this dataset with low frequency in our simulation. We report the results on the validation set in Table.~\ref{tab:test on nu}.
It can observed that compared with PolarNet backbone, these two proposed balanced modules achieve the consistent performance gain. For these tailed categories, including bicycle, motorcycle, trailer and traffic cone, a significant improvement is achieved by adding these two modules. The cooperation  of these two modules leads to the proposed input-output balanced model, which achieves the superior performance.

\subsection{Ablation Studies}
In this section, we perform the ablation experiments to verify the effect of different components and investigate the generalization ability of the proposed modules.

\noindent\textbf{Effects of Components}
The results on SemanticKITTI validation set are reported in Table. \ref{tab:Ablation1}. The baseline is the PolarNet. It can be observed that both balanced modules contribute an improvement compared to the baseline, which indicates the effects of input data balance and output balance.

\noindent\textbf{Experiment on Different Backbones}
In this experiment, we use two different backbones to verify the generalization ability of the input balanced module. These backbones are PolarNet and SalsaNext. SalsaNext~\cite{cortinhal2020salsanext} is built upon the base SalsaNet~\cite{cortinhal2020salsanext} model which utilize the spherical projection with encoder-decoder structure. As shown in Table. \ref{tab:Ablation2}, it can be observed that the input balanced module delivers consistent performance improvement on both backbones, \ie, about $2\%$ mIOU gain on PolarNet and $1\%$ mIOU gain on the better backbone SalsaNext.

\section{Conclusion}

In this paper, we focus on the long-tailed issue in LiDAR semantic segmentation task and have proposed an input-output balanced framework to handle this inherent difficulty. Specifically, we first employ the mesh models to create the point cloud of these tail categories, which better simulates the position and density of original dataset and further enhance the data diversity. Then, an output balanced module consisting of multi-head block, is designed to alleviate the interference of dominating categories and promote the feature learning for tail categories. Extensive experiments on two large-scale datasets demonstrate the effectiveness of the proposed input-output balanced model. Ablation studies also show that these balanced modules have good generalization ability when applying to different backbones.

\bibliographystyle{IEEEbib}
\bibliography{icme2021template}

\end{document}